\documentclass[conference]{IEEEtran}
\IEEEoverridecommandlockouts
\usepackage{cite}
\usepackage{amsmath,amssymb,amsfonts}
\usepackage{algorithmic}
\usepackage{graphicx}
\usepackage{textcomp}
\usepackage{xcolor}
\usepackage{url}
\usepackage[ruled, linesnumbered]{algorithm2e}
\def\BibTeX{{\rm B\kern-.05em{\sc i\kern-.025em b}\kern-.08em
    T\kern-.1667em\lower.7ex\hbox{E}\kern-.125emX}}
\begin{document}

\title{BLoad: Enhancing Neural Network Training with Efficient Sequential Data Handling\\
}

\author{A S M~Iftekhar\textsuperscript{*},
        Raphael~Ruschel\textsuperscript{*},
        Suya You, 
        B. S. Manjunath
\thanks{* represents equal contribution.}
\thanks{A S M Iftekhar, Raphael Ruschel, and B.S. Manjunath are with the Department
of Electrical and Computer Engineering, University of California Santa Barbara, Santa Barbara,
CA, 93106. E-mail: \{iftekhar, raphael251, manj\}@ucsb.edu}
\thanks {Suya You is with Army Research Laboratory, Intelligent Perception, CISD
Los Angeles, CA. Email: suya.you.civ@army.mil}

}

\maketitle

\begin{abstract}
The increasing complexity of modern deep neural network models and the expanding sizes of datasets necessitate the development of optimized and scalable training methods. In this white paper, we addressed the challenge of efficiently training neural network models using sequences of varying sizes. To address this challenge, we propose a novel training scheme that enables efficient distributed data-parallel training on sequences of different sizes with minimal overhead. By using this scheme we were able to reduce the padding amount by more than 100$x$ while not deleting a single frame, resulting in an overall increased performance on both training time and Recall in our experiments. 
\end{abstract}


\section{Introduction}

The increasing complexity of modern deep neural network models and the expanding sizes of datasets necessitate the development of optimized and scalable training methods. Neural networks are commonly trained using multiple GPUs either within a single machine or distributed across a cluster of nodes. Traditional distributed training schemes, such as distributed data-parallel (DDP)\cite{torchDataParallel}, have been widely employed. While this scheme is popular, it struggles with data sequences of varied lengths, like videos of different durations. To address this challenge, we propose a novel training scheme that enables efficient DDP training on sequences of different sizes with minimal overhead and is publicly available at \textcolor{blue}{\url{https://github.com/RRuschel/BLoad}}.

\section{Problem Statement and Current Limitations}

We consider a dataset $\mathcal{D}$ comprising $N$ samples, where each sample $S_{i \in N}$ represents a video with dimensions $H \times W \times T$. Here, $H$ and $W$ denote the height and width of each frame, respectively, while $T$ represents the duration of the video. Our objective is to train a deep neural network model efficiently using a DDP scheme while accommodating varying values of $H$, $W$, and $T$ for each sample $S_i$. While our primary focus is on videos, we expect our method to be applicable to other data types like audio and text.

Using PyTorch's Distributed Data-Parallel with datasets of varying lengths can lead to stalled training without any error message. The root of this issue is in the gradient synchronization step. Here, each GPU process collects gradients from other processes to compute an average gradient, which updates the model. If sequences differ in size, each process gets varying sample counts, potentially causing a deadlock as processes await others indefinitely, unable to calculate the gradient.

\begin{figure}
    \centering
    \includegraphics[width=\linewidth]{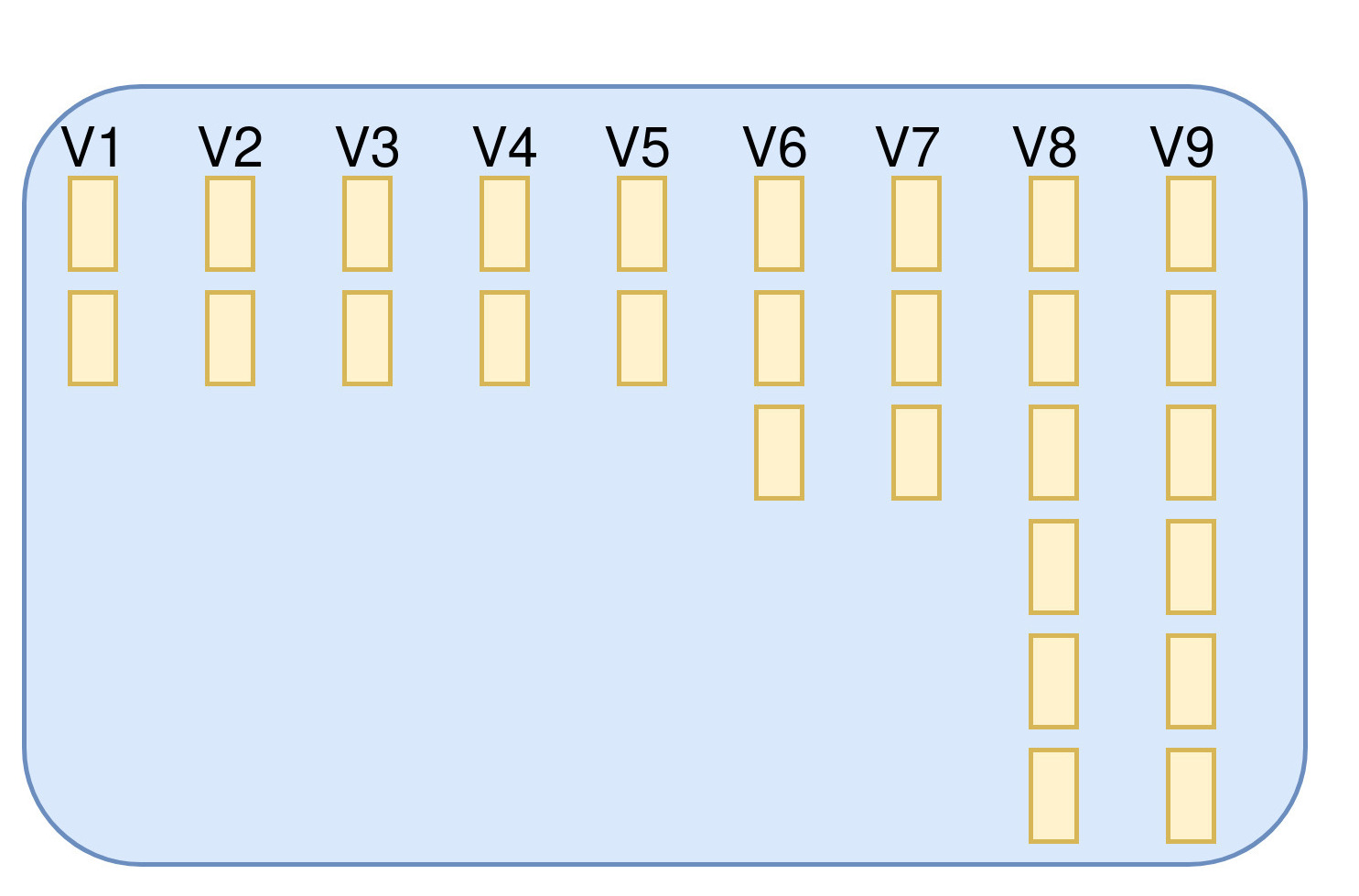}
    \caption{Sample dataset with 8 videos of varying length - Each $V_i$ represents an individual video, and each yellow square represents a frame.}
    \label{fig:dataset}
\end{figure}

\begin{figure}[h]
    \centering
    \includegraphics[width=0.8\linewidth]{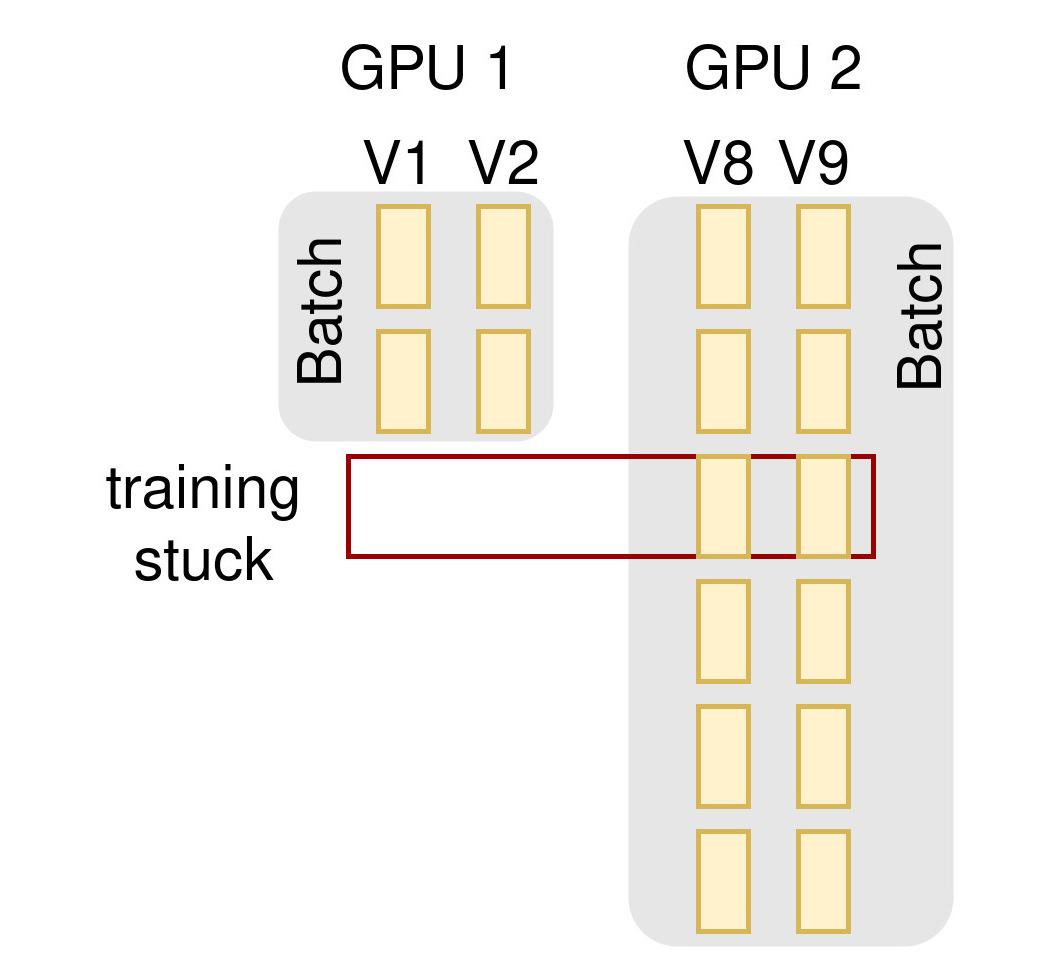}
    \caption{Deadlock situation when each GPU receives sequences of different lengths. In this situation, after the third iteration, GPU 1 will not have any gradients to report, causing GPU 2 to wait without any error message.}
    \label{fig:deadlock}
\end{figure}

To illustrate the problem, consider the sample dataset from figure \ref{fig:dataset}. It has 8 sequences with lengths varying from 2 to 6 frames. Initiating a DDP training with a batch size of 2 using random sampling can produce situations like that in figure \ref{fig:deadlock}. Here, GPU 1 handles two videos, each 2 frames long, while GPU 2 manages two videos, each 6 frames long. After just 2 iterations, GPU 1 completes its batch, leaving it idle, as GPU 2 continues processing. New data is only retrieved once all GPUs finish their batches. Consequently, when GPU 2 tries to gather gradients from GPU 1, it faces an indefinite wait since GPU 1 has no gradient to return.

\begin{figure}
    \centering
    \includegraphics[width=0.8\linewidth]{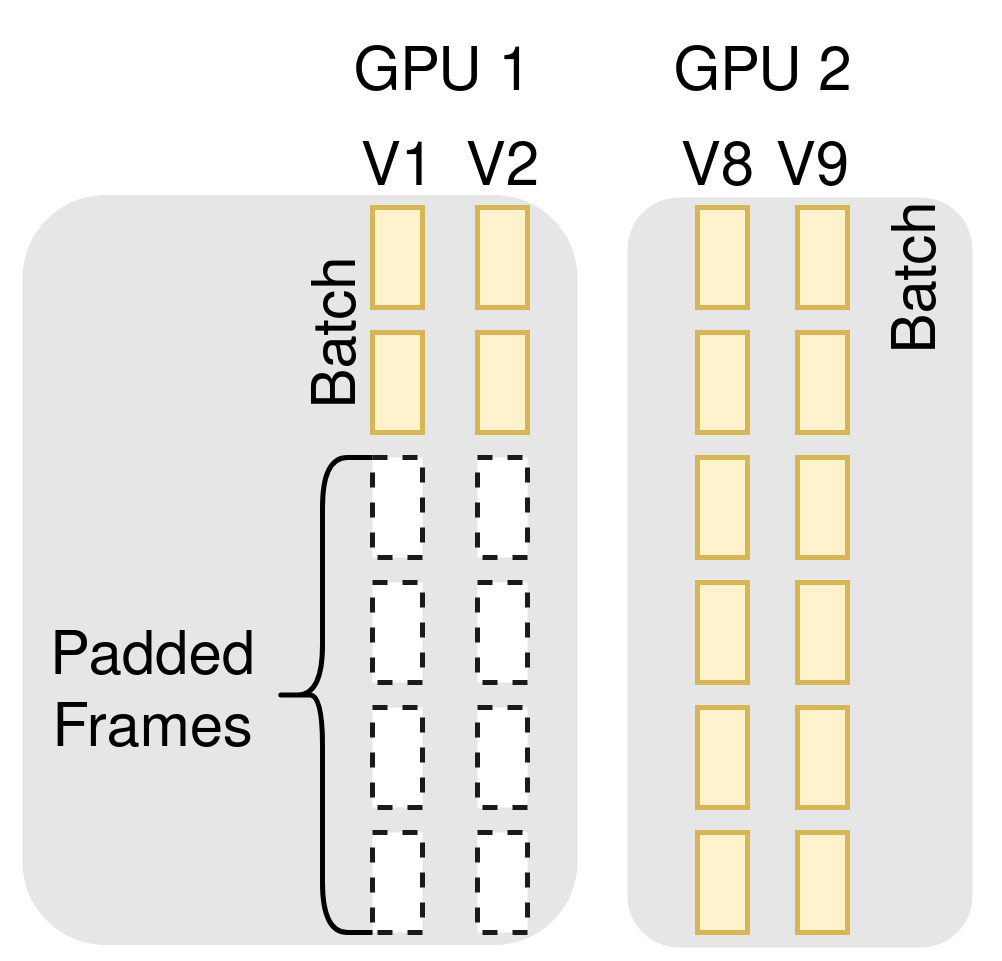}
    \caption{Naive padding solution - Every sequence on the dataset is padded to match the length of the largest sequence, generally by adding $0$'s or repeating the last entry of the sequence}
    \label{fig:pad}
\end{figure}

A common strategy to resolve this issue involves padding each sample to match the duration $T_{max}$ of the longest sequence in the dataset (as illustrated in figure \ref{fig:pad}). While this method solves the stalling problem, it becomes highly inefficient when $T_{max}$ is significantly larger than the average sequence length, resulting in substantial padding and unnecessary computations during training.

\begin{figure}
    \centering
    \includegraphics[width=\linewidth]{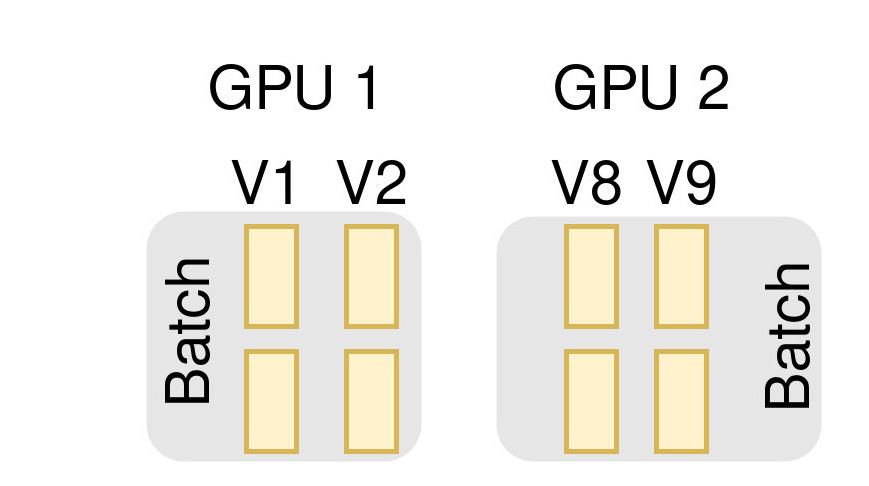}
    \caption{Sampling solution, where each sequence is trimmed to match a smaller size, usually the length of the average entry in the dataset. In this approach, one sequence might be broken into several smaller portions, which won't allow the training of models with long temporal support.}
    \label{fig:sample}
\end{figure}

Another strategy entails breaking down each data sample into smaller chunks of size $H \times W \times T_{block}$ and treating each smaller block as an individual sample, as employed in \cite{zeng2022motr}, \cite{meinhardt2022trackformer}. While this approach resolves synchronization issues, it cannot be applied to train neural networks that incorporate feedback, such DDS\cite{iftekhar2023dds}. Breaking down the original data sample into smaller pieces destroys the temporal relationships inherent in the original sequence, as shown in figure \ref{fig:sample}.

\section{Methods}

\begin{figure}
    \centering
    \includegraphics[width=\linewidth]{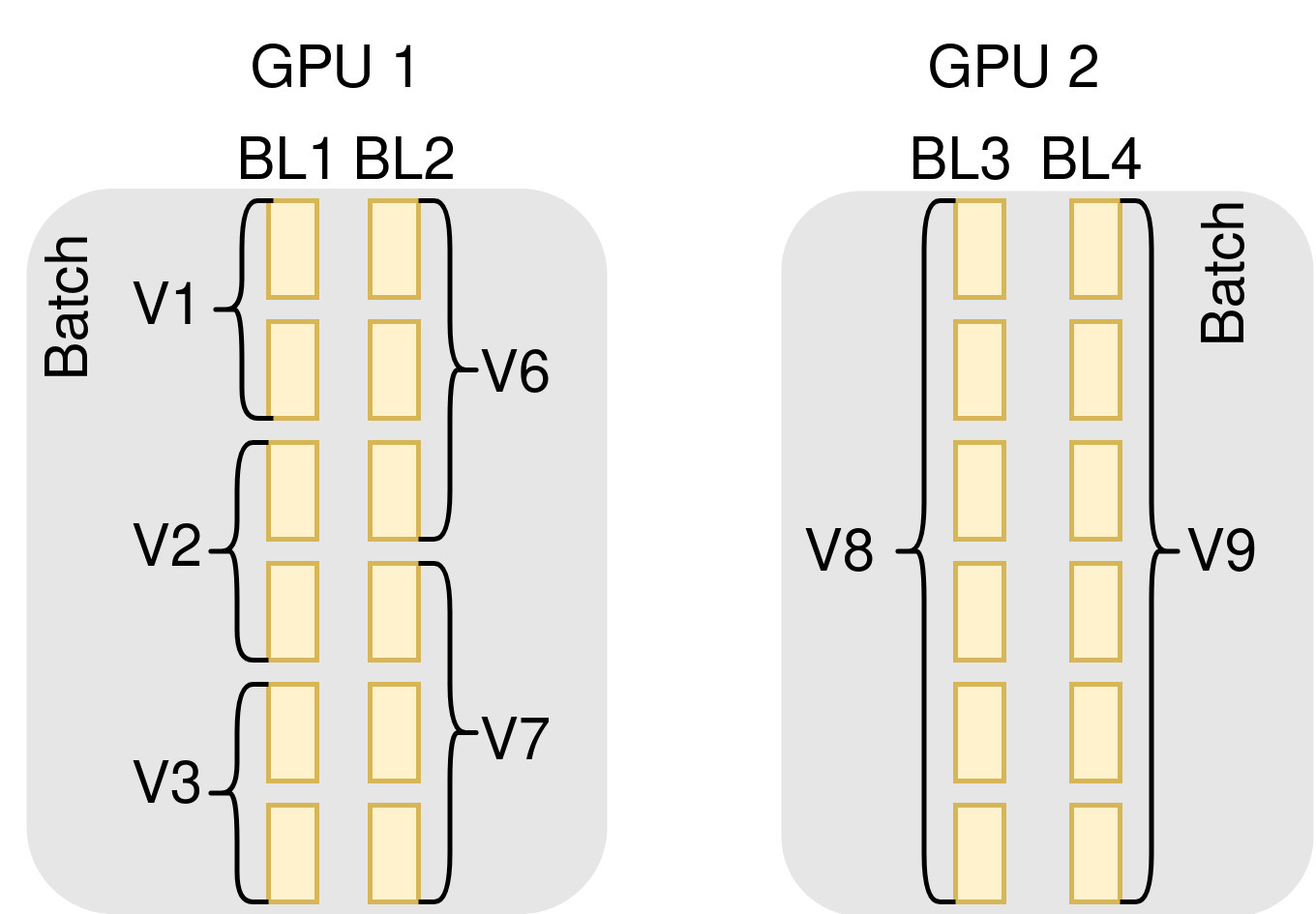}
    \caption{Our proposed padding approach - \textbf{BLoad} (as in block load) - aims to construct sequences of size $T_{max}$ using shorter sequences as building blocks}
    \label{fig:solution}
\end{figure}

Our proposed method builds upon the padding strategy but significantly reduces wasteful computations. We create blocks of size $T_{max}$ by concatenating randomly sampled sequences with length $T_i \le T_{max}$. If we cannot construct a video of size exactly equal to $T_{max}$, we then build a block as close to $T_{max}$ and then pad it with $0$'s to fill the block. Figure \ref{fig:solution} shows an example of our proposed solution.  

\begin{figure}
    \centering
    \includegraphics[width=\linewidth]{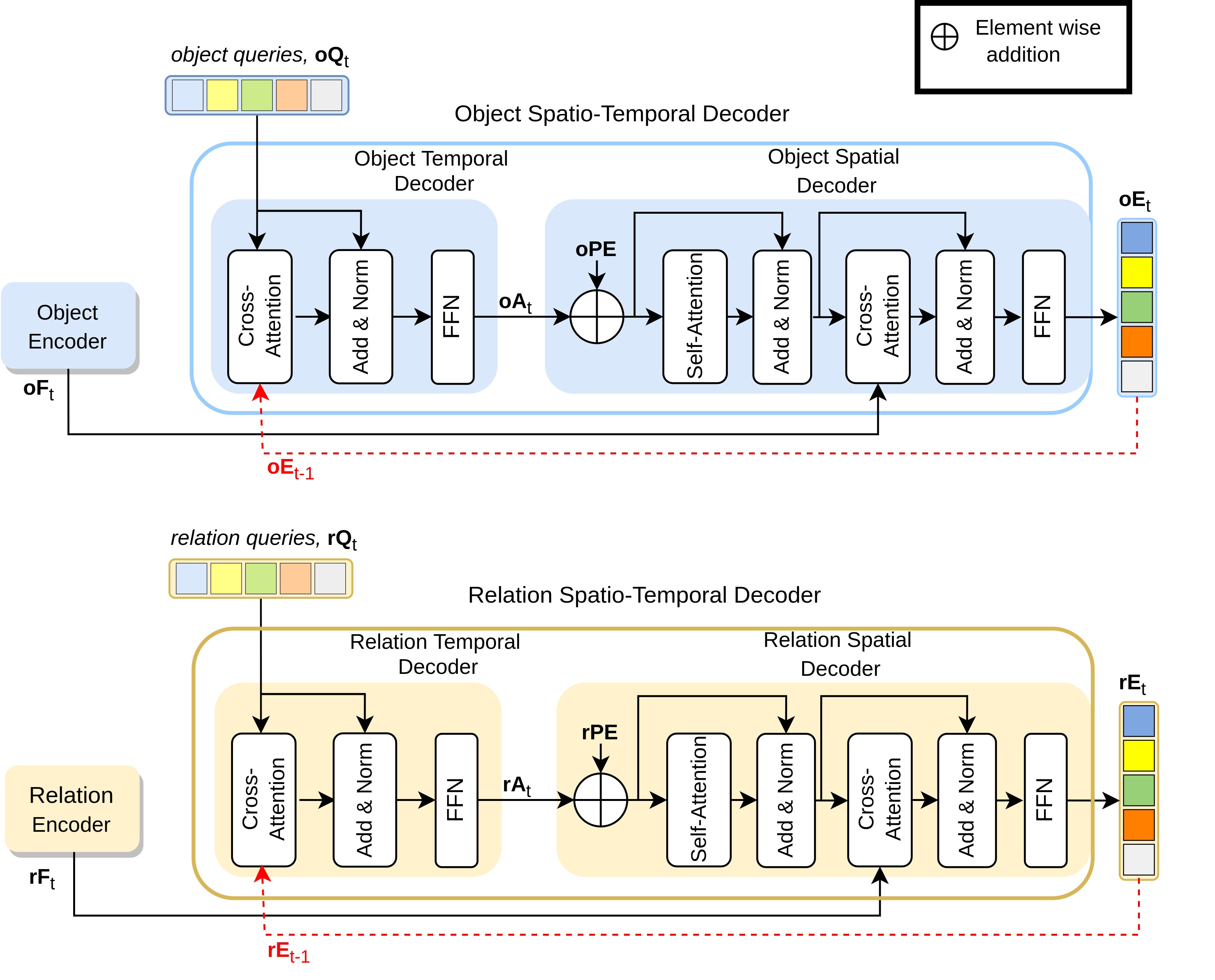}
    \caption{DDS architecture from \cite{iftekhar2023dds} - In this model, the output of frame $n-1$ is used as partial input to both encoders during the processing of frame $n$, resulting in increased performance on video sequences.}
    \label{fig:DDS}
\end{figure}

Additionally, we create a table containing the starting index of each new video within each particular block. This table can be useful during training a recurrent network, such as the DDS, architecture shown in Figure \ref{fig:DDS} where some information ($oE_{t-1}$) from iteration $n-1$ is used at iteration $n$. Having the knowledge of where a new sequence starts enables resetting/discarding the information from the previous iteration, as it belongs to a different sequence, correctly maintaining the temporal dependency of the data inside each block.

\begin{figure}
    \centering
    \includegraphics[width=\linewidth]{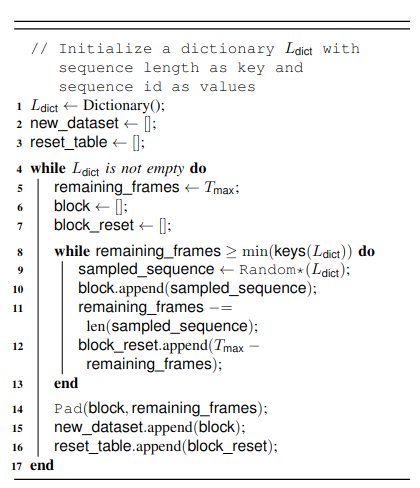}
    \caption{Pseudocode for our proposed padding approach - Note that the $Random_*()$ function returns a random entry of $L_{dict}$ such that $T_{sampled\_sequence} \leq remaining\_frames$}
    \label{fig:algo}
\end{figure}

For a more technical insight into our method, we've provided a pseudocode outline on \ref{fig:algo}.

\section{Experiments \& Results}


Following the works on \cite{iftekhar2023dds}, we perform experiments on the Action Genome dataset. This dataset is extensively used in Scene-Graph Detection problems and contains $7,464$ videos with $166,785$ frames in the training set and $1,737$ videos with $54,371$ frames in the test set. To evaluate the differences between each sampling strategy, we retrain DDS using each strategy mentioned earlier and report the amount of padding added, number of frames deleted, time per epoch, and performance on the recall@20 metric. The Action Genome dataset is apt for these experiments, given its wide range of sequence lengths, from brief 3-frame snippets to as long as 94 frames. Our experiments were conducted on a machine with 8 NVIDIA A100 with 40GB of memory and are reported in table \ref{tab:results}. 

\begin{table}[htbp]
\centering
\caption{Comparison of performance using different training strategies}
\begin{tabular}{lcccc}
\hline
 & \textbf{0 padding} & \textbf{sampling} & \textbf{mix pad} & \textbf{block\_pad} \\
\hline
\textbf{padding amount} & 534831 & 0 & 37712 & 3695 \\

\textbf{\# frames deleted} & 0 & 92271 & 40289 & 0 \\

\textbf{time (per epoch)} & 170 min & 18 min & 40 min & 41 min \\

\textbf{recall@20} & - & 41.2 & 42.1 & 43.3 \\
\hline
\end{tabular}
\label{tab:results}
\end{table}

From the table, it's evident that the naive padding solution results in over $500k$ padding frames—almost $4x$ the original data size. This rendered the training so inefficient that we chose not to complete it for performance evaluation. Interestingly, with the sampling strategy, despite discarding nearly $2/3$ of the data, we achieved results comparable to or even surpassing several established models such as \cite{cong2021spatial}. We attribute this to the dataset's high frame correlation, leading to marginal gains with added frames. We haven't delved deeper into this observation as it falls outside this manuscript's scope. Our proposed block pad strategy offers clear advantages. It combines zero frame removal with minimal padding, reducing unnecessary computations and enhancing performance.

\section{Conclusion}

In this white paper, we addressed the challenge of efficiently training neural network models using sequences of varying sizes. We proposed a novel training scheme that combines elements of padding and distributed data parallelism to achieve optimal results. By padding sequences with videos of appropriate lengths and employing a table of starting indices, our method reduces wasteful computations while preserving temporal relationships. The proposed approach opens up new possibilities for training models on diverse data types, such as videos, audio, and text, with varying sequence lengths. In future research, we can delve into the method's applicability to different modalities and test its efficacy across various deep learning challenges.

\bibliography{conference_101719}
\bibliographystyle{IEEEtran}

\end{document}